\documentclass[letterpaper, 10 pt, conference]{ieeeconf}  

\IEEEoverridecommandlockouts                              

\usepackage{graphics} 
\usepackage{epsfig} 
\usepackage{mathptmx} 
\usepackage{times} 
\usepackage{amsmath} 
\usepackage{amssymb}  
\usepackage{graphicx}
\usepackage[OT1]{fontenc}
\newcommand\shift{-2mm}
\usepackage[colorlinks=true, linkcolor=black, urlcolor=black]{hyperref}

\title{\LARGE \bf
Vertical Vibratory Transport of Grasped Parts Using Impacts}

\author{C. L. Yako$^{\dagger,*}$, J\'er\^ome Nowak$^{*}$, Shenli Yuan, and Kenneth Salisbury
\thanks{$\dagger$ Corresponding author}
\thanks{*These authors contributed equally to this work.}
\thanks{Connor Yako and J\'er\^ome Nowak are with the Department of Mechanical Engineering, Stanford University, Stanford, CA 94305 USA (email: \{connor.yako, jerome.nowak\}@stanford.edu). Shenli Yuan is with SRI International, Menlo Park, CA 94025 USA (email: shenli.yuan@sri.com). Kenneth Salisbury is with Department of Computer Science, Stanford University, Stanford, CA 94305 USA (email: kenneth.salisbury@gmail.com).}%
}

\begin{document}

\maketitle
\thispagestyle{empty}
\pagestyle{empty}

\begin{abstract}

In this paper, we use impact-induced acceleration in conjunction with periodic stick-slip to successfully and quickly transport parts vertically against gravity. We show analytically that vertical vibratory transport is more difficult than its horizontal counterpart, and provide guidelines for achieving optimal vertical vibratory transport of a part. Namely, such a system must be capable of quickly realizing high accelerations, as well as supply normal forces at least several times that required for static equilibrium. We also show that for a given maximum acceleration, there is an optimal normal force for transport. To test our analytical guidelines, we built a vibrating surface using flexures and a voice coil actuator that can accelerate a magnetic ram into various materials to generate impacts. The surface was used to transport a part against gravity. Experimentally obtained motion tracking data confirmed the theoretical model. A series of grasping tests with a vibrating-surface equipped parallel jaw gripper confirmed the design guidelines. The code is public on \href{https://github.com/clyako/vertical-vibratory-transport-of-grasped-parts-using-impacts.git}{\underline{GitHub}} and the video presentation is on \href{https://youtu.be/Mb02fUOyaTE}{\underline{YouTube}}. 

\end{abstract}

\section{INTRODUCTION}\label{section: introduction}



Using vibrations to manipulate parts is not a new concept and has been used for non-prehensile motion for decades. 
Popular implementations include vibratory part feeders, where parts experience sequential sticking, slipping, non-contact ``hopping'', and landing phases caused by a combination of in and out-of-plane vibrations of the drive surface~\cite{lim1997conveying}. 
However, this ``hopping'' phase may be unsuitable for precise in-hand manipulation tasks. 
This limitation led Reznik and Canny to develop their ``Coulomb Pump'' based on only 1D vibrations and an always sliding assumption~\cite{reznik1998coulomb}. 
Parts moved due to \textit{time-asymmetry}, where the drive surface spent more time moving in the desired transport direction across one oscillation cycle.
Quaid built upon the ``Coulomb Pump'' by removing the always sliding assumption and noting that part transport could be achieved by a simple two-phase cycle: a slow \textit{sticking} phase in which the surface and part moved in the desired motion direction, followed by a high backward acceleration \textit{slipping} phase where the surface slips behind the part~\cite{quaid1999feeder}.
Umbanhowar and Lynch took Quaid's work further by deriving acceleration profiles that maximized a part's average velocity over one cycle~\cite{umbanhowar2008optimal}.
Instead of Quaid's two phases, Umbanhowar and Lynch added a third phase where the surface caught up to the part before its velocity became negative.
They also described how part speed could be improved by simultaneously oscillating in the vertical direction with the constraint that contact be maintained.

%
\begin{figure}[t]
    \centering
    \includegraphics[width=0.44\textwidth]{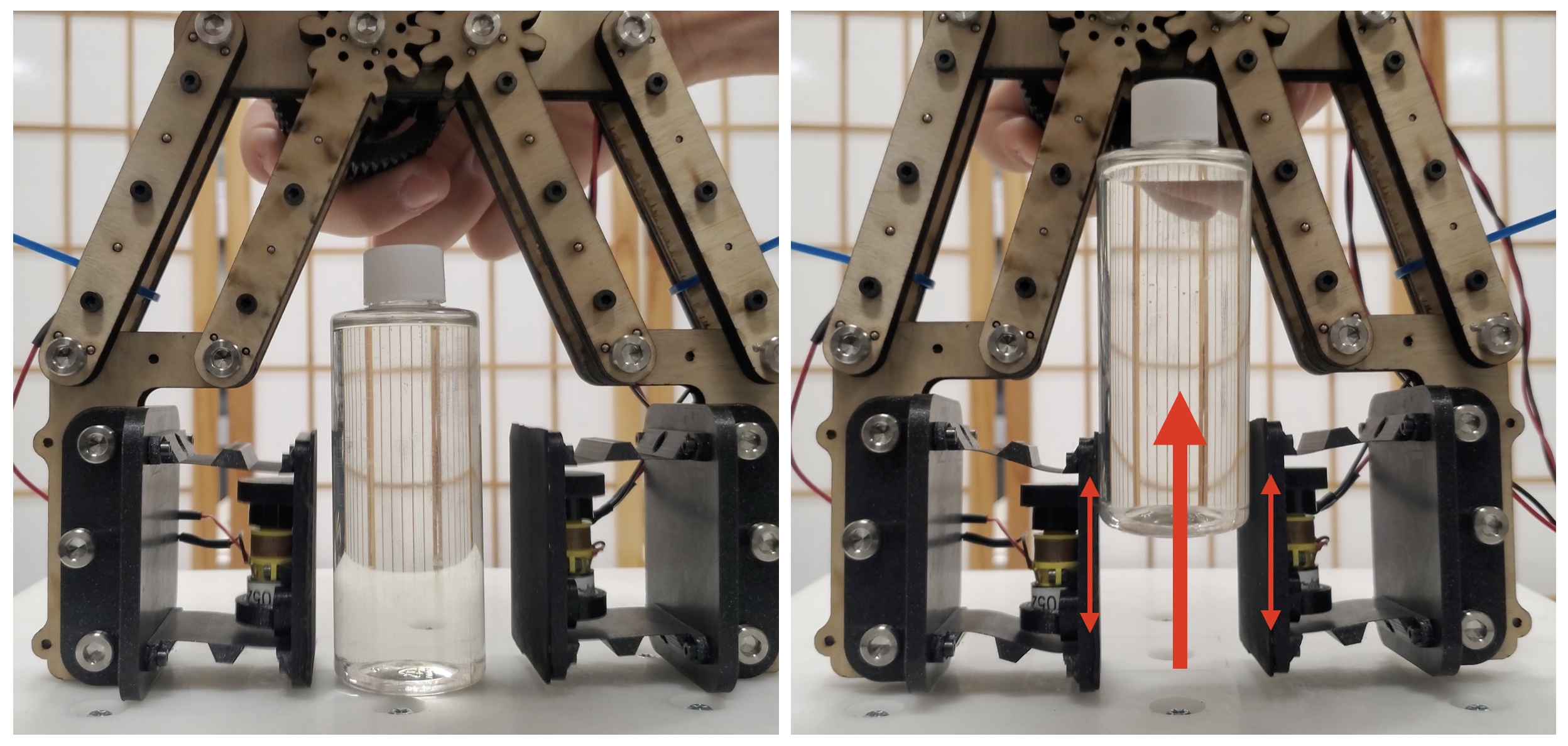}
    \caption{Vibrating surfaces lift a part against gravity.}
    \label{figure: cover photo}
    \vspace{\shift}
\end{figure}
%

A recent paper by Nahum used this idea by cleverly leveraging the smoothly changing in/out-of-plane vibration direction of an eccentric rotating mass (ERM) motor in conjunction with a grasp on the part to enforce contact~\cite{nahum2022robotic}. 
Thus, the rotating vibration direction changed the normal force, and hence the friction force, making it easier to transition from sticking to slipping as well as stay in the sticking phase. 
The authors noted that their device could work against gravity up to 60$^\circ$ from the horizontal, and that a stronger motor would be needed above this inclination.

\subsection{Upward Vertical Vibratory Transport is Difficult}

For simplicity's sake, this paper will focus on moving a part vertically against gravity using a 1D vibratory surface, with the vibrations in the direction of part movement.
Given a desired part motion, vertical transport is inherently more difficult than its horizontal counterpart as gravity biases the friction cone such that it is easier for the part to slip downwards rather than upwards. 
Take the two phases described by Quaid, the \textit{sticking} phase and the \textit{slipping} phase. 
This skewed friction cone limits the peak upwards surface acceleration during sticking and increases the necessary downward surface acceleration to transition to slipping.
Gravity compounds the part's downward acceleration during slipping.

Despite these difficulties, we show that it is possible to achieve purely vertical transport, i.e., against gravity, using only 1D vibrations (Fig. \ref{figure: cover photo}). 
This paper is organized as follows. 
In Section \ref{section: dynamics} we give the equations of motion for vertical vibratory transport, we show why upward motion is difficult to achieve, which conditions need to be met and why we chose to use rigid body impacts to create high accelerations.
We also derive the optimal vibration waveform by showing an equivalence with horizontal transport.
Section \ref{section: design of experimental setup} describes the design of a device capable of achieving vertical transport.
Section \ref{section: model validation} describes the experimental setup used to validate our dynamical model.
A gripper design is presented and preliminary grasping transport tests are reported in Section \ref{section: gripper}.
We then conclude with a comparison of our device  with direct drive devices and limitations in Section \ref{section: discussion}.

\section{DYNAMICAL MODEL}\label{section: dynamics}



\subsection{Conditions Necessary To Achieve Upward Transport}

\begin{figure}[t]
    \centering
    \includegraphics[width=0.45\textwidth]{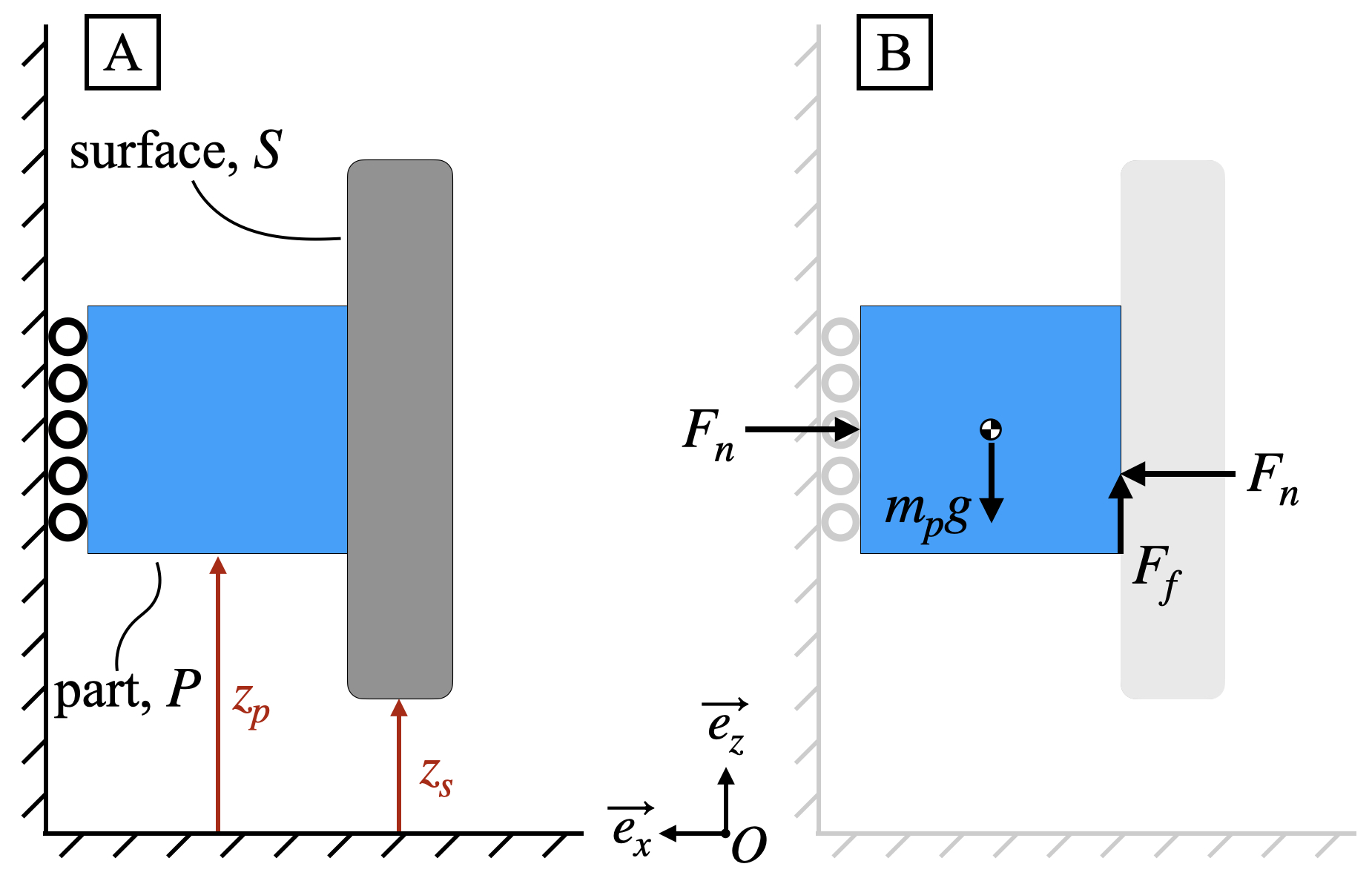}
    \caption{Kinematics and dynamics.
    The part, $P$, and the surface, $S$ are constrained to move purely along the $\Vec{e_z}$ direction.
    The rollers on the left side of $P$ indicate sliding on a frictionless surface.
    %
    }
    \label{figure: kinematics and dynamics}
    \vspace{\shift}
\end{figure}

Our notations follow the conventions in Umbanhowar~\cite{umbanhowar2008optimal}. The part $P$ of mass $m_P$ is subject to its weight $-m_P g \Vec{e_z}$ and is held between a vertical vibrating surface $S$ with normal force $F_n \Vec{e_x}$ and tangential friction force $F_f \Vec{e_z}$ on one side, and a frictionless vertical support on the other side.
We assume dry Coulomb friction between $P$ and $S$ with coefficients of static and kinetic friction  $\mu_s \geq \mu_k$, respectively.
Let $z_S$ be the vertical position of the vibrating surface with respect to a stationary point $O$ and $z_P$ that of the part (Fig. \ref{figure: kinematics and dynamics}). We seek mechanical conditions and surface motion which will transport the part upward over time.

The equation of motion of the part is: $m_P \ddot{z}_P = F_f - m_P g$. If the part is initially stationary with respect to the surface, the part sticks to the surface ($\dot{z}_P = \dot{z}_S$) as long as $|F_f| = m_P|\ddot{z}_S + g| \leq \mu_s F_n$.
Therefore, to prevent the part from slipping down when the surface is stationary ($\dot{z}_S = 0$), we need:
\begin{equation}\label{eq: F_n lower bound mu_s}
    F_n > m_P g / \mu_s
\end{equation}
At the onset of slipping when $m_P|\ddot{z}_S + g| > \mu_s F_n$, the friction force is $F_f = \mu_k F_n \text{sgn}(\ddot{z}_S + g)$.
If the part is already slipping $(\dot{z}_P \neq \dot{z}_S)$, then $F_f = \mu_k F_n \text{sgn}(\dot{z_S} - \dot{z}_P)$ until the conditions $\dot{z_P} = \dot{z}_S$ and $|F_f| \leq \mu_s F_n$ are reached.
In summary, the equations of motion are:
\begin{flalign}
    \label{eq:EOM sticking}
    \quad \textit{sticking:} \quad & \dot{z}_P = \dot{z}_S, \quad -\frac{\mu_s F_n}{m_P} -g \leq \ddot{z}_S \leq \frac{\mu_s F_n}{m_P} - g&&\\
    \label{eq:EOM slipping}
    \quad \textit{slipping:} \quad & \ddot{z}_P = \frac{\mu_k F_n}{m_P} \text{sgn}(\dot{z}_S - \dot{z}_P) - g, \quad \dot{z}_P \neq \dot{z}_S&&
\end{flalign}
 To prevent the part from falling down indefinitely during slipping, \eqref{eq:EOM slipping} yields the requirement:
\begin{equation}\label{eq: F_n lower bound mu_k}
    F_n > m_P g / \mu_k
\end{equation}

Let the surface oscillate vertically with period $T$. 
We assume we can directly prescribe the surface motion up to a maximum acceleration $|\ddot{z}_S| \leq a_{max}$ corresponding to the physical limits of the surface actuator.
As a periodic function of time, $\dot{z}_S$ is bounded, thus the part velocity is bounded during sticking. 
%
%
During slipping, from \eqref{eq:EOM slipping}, \eqref{eq: F_n lower bound mu_k}, and periodic $z_S$ the part has finite acceleration/deceleration phases, bounding $\dot{z}_P$ at all times.
Therefore, excluding edge cases, starting from rest the part reaches a steady state where its velocity is $T$-periodic with a constant average velocity: $v_{ave} = (z_P(t+T) - z_P(t))/T$. 
Because the surface has no net travel, upward part motion ($v_{ave} > 0$) requires the surface to slip down with respect to the part, so \eqref{eq: F_n lower bound mu_s} and \eqref{eq:EOM sticking} yield:
\begin{equation}\label{eq: a_max lower bound}
    a_{max} > \mu_s F_n/m_P + g > 2g
\end{equation}

Equations \eqref{eq: F_n lower bound mu_s} to \eqref{eq: a_max lower bound} quantify the challenges which upward vertical transport presents compared to horizontal transport.
In most practical cases, we have $\mu_k < \mu_s < 1$, so \eqref{eq: F_n lower bound mu_s} and \eqref{eq: F_n lower bound mu_k} mean that the part $S$ must be squeezed with a force $F_n$ exceeding its own weight.
Equation \eqref{eq:EOM sticking} shows that gravity reduces the maximum upward part acceleration during sticking, and that overcoming this limitation requires squeezing the part harder still.
However, from \eqref{eq: a_max lower bound}, squeezing with higher normal forces requires more powerful actuators to reach higher surface accelerations, which already need to exceed $2g$ (compared to $a_{max} > \mu_s g$ for the horizontal case).
Finally, equation \eqref{eq:EOM slipping} shows that during slipping, the part accelerates faster down than up.

Note that replacing the frictionless support with a second, synchronized vibrating surface doubles $F_f$, yielding the same equations of motion but instead with $m_P / 2$: for a given $F_n$, parts twice as heavy can be lifted. 
%
%
This can be visualized by mirroring the part, surface, and rollers in Fig.~\ref{figure: kinematics and dynamics}B about the vertical wall to the left of the rollers; there is $2m_p$ between the two synchronized surfaces and the inside surface of each part experiences $F_n$ transmitted through the wall.
While we assume completely in-phase surfaces in this work, out-of-phase surfaces could be a topic of future investigation.
Furthermore, the limit where $F_n \gg m_P g$ while still satisfying \eqref{eq: a_max lower bound} is equivalent to horizontal transport of a part of mass $m_P$ with acceleration of gravity equal to $F_n/m_P$. 

\subsection{Optimal Upward Stick-Slip Transport}\label{section: optimal upward}

Our analysis adapts the optimization theory for horizontal stick-slip transport presented by Umbanhowar and Lynch~\cite{umbanhowar2008optimal}. For completeness, we reproduce their logic here as applied to our case; namely, we add gravity in the plane of motion and make the normal force $F_n$ an extra parameter.

We would like to find the periodic surface acceleration $\ddot{z}_S$ which maximises the average part velocity $v_{ave}$. Formally, given a period $T$ and a maximum surface acceleration $a_{max}$, find $\ddot{z}_S \colon [0,T] \rightarrow \left[a_{max}, a_{max}\right]$ satisfying the continuity conditions $z_S(0) = z_S(T)$ and $\dot{z}_S(0) = \dot{z}_S(T)$ which maximises:
\begin{equation}\label{eq: maximize L}
    \int_0^T L(t) dt = \int_0^T (\dot{z}_P(t)-\dot{z}_S(t)) dt
\end{equation}

Umbanhowar and Lynch make the two following observations, which still hold in our case.

\textit{Observation 1:} ``During any periodic steady-state motion, there
exists a $t \in \left[0,T\right]$ such that $\dot{z}_S(t) = \dot{z}_P(t)$.'' 
Indeed, if this were not the case, then the part would always be slipping and since the velocities are continuous functions, the slipping would occur in a fixed direction. 
From \eqref{eq:EOM slipping}, $\ddot{z}_P$ would therefore have a constant non-zero value, which is incompatible with steady-state velocity.

\textit{Observation 2:} ``For any optimal solution, $\dot{z}_S(t) \leq \dot{z}_P(t)$ for all $t \in \left[0,T\right]$.'' 
This observation is more subtle than the previous one and follows from $\mu_s > \mu_k$. 
Indeed, \textit{Observation 1} guarantees that for any interval where the part is slipping down relative to the surface ($\dot{z}_S(t) > \dot{z}_P(t)$), there exists an onset $t_0$ just before which the part was sticking to the surface. 
For $t \geq t_0$ and throughout that slipping period, the same $\dot{z}_P(t)$ can be achieved through sticking with $\dot{z}_S(t) = \dot{z}_P(t)$ while keeping $\dot{z}_S$ continuous, which leads to a larger value of $L$ and a more optimal solution.

\begin{figure}[t]
    \centering
    \includegraphics[width=0.45\textwidth]{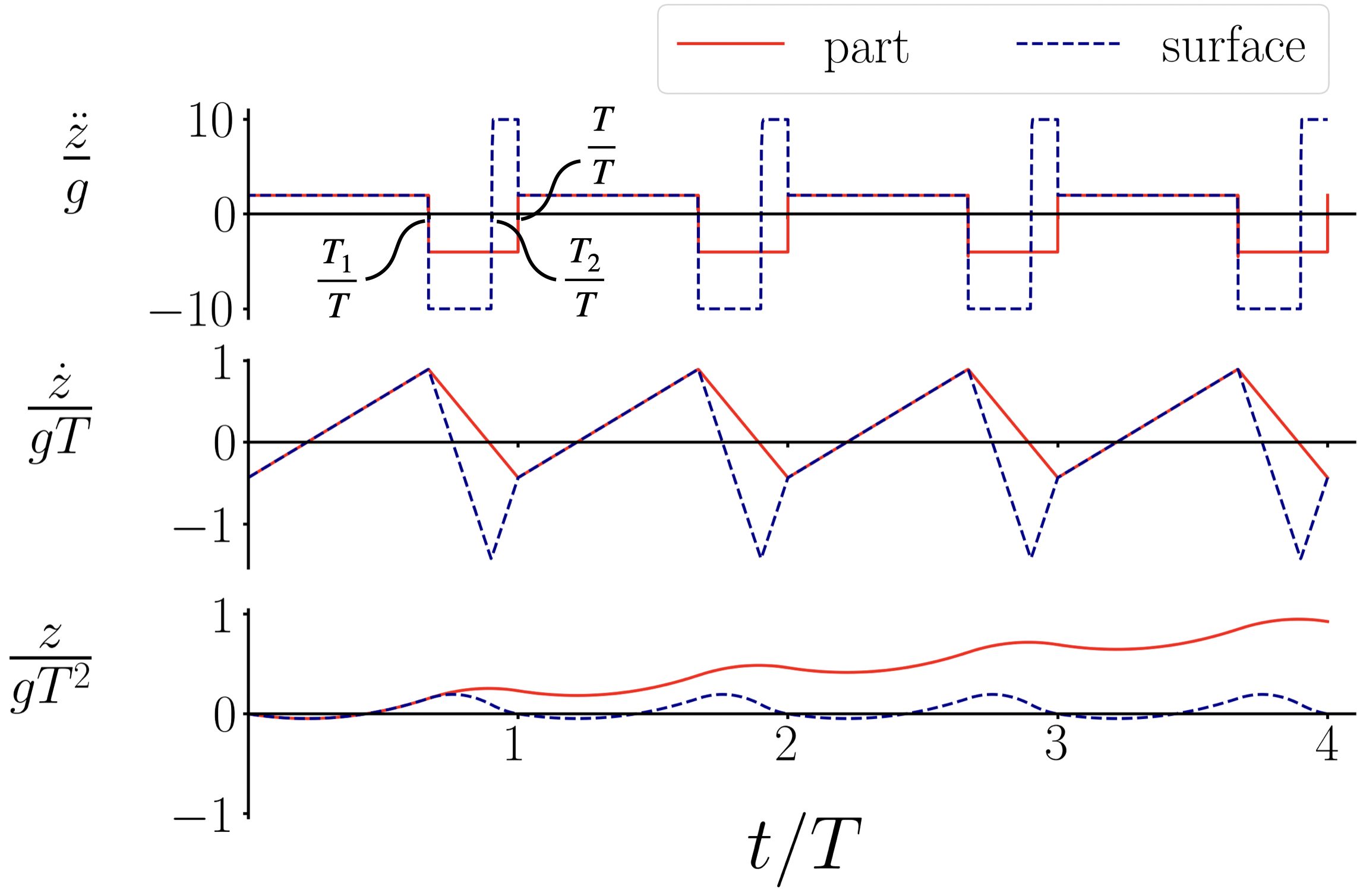}
    \caption{Nondimensional accelerations, velocities, and positions of the surface (blue) and part (red) given $f_n = 5$ and $a_{max} = 10g$. Despite the surface achieving $10g$ of acceleration, which is used to slip below and catch up to the part, the part's velocity does go negative unlike in previous work~\cite{umbanhowar2008optimal}.}
    \label{figure: optimal motion}
    \vspace{\shift}
\end{figure}

These observations show that for an optimal surface motion, the part either sticks to the surface or slips up relative to it. 
Therefore, the only static friction limit which is crossed in \eqref{eq:EOM sticking} is $\ddot{z}_S < -\mu_s F_n/m_P -g$ and the only value attained by $\ddot{z}_P$ in \eqref{eq:EOM slipping} is $\ddot{z}_P = -\mu_k F_n/m_P - g$.
So, although gravity makes the equations of motion \eqref{eq:EOM sticking} and \eqref{eq:EOM slipping} asymmetric in $z \leftrightarrow -z$, this dynamical asymmetry is never encountered during optimal surface motion.
Crucially, this means that the optimization results in Umbanhowar and Lynch~\cite{umbanhowar2008optimal} for horizontal equations of motion, which are symmetric in $x \leftrightarrow -x$, can also be applied to vertical motion despite the asymmetry in $z$.
In fact, as long as we don't allow the part to slip down relative to the surface (\textit{Observation 2}), the equations of motion for upward vertical transport given $\mu_s$ and $\mu_k$ are the same as the equations of motion for horizontal transport given fictional friction coefficients $\widetilde{\mu_s} = \mu_s F_n/(m_P g) - 1$ and $\widetilde{\mu_k} = \mu_k F_n/(m_P g) + 1$.
Note the peculiarity $\widetilde{\mu_s} < \widetilde{\mu_k}$, which theoretically means having the part slip down relative to the surface would allow greater upward part accelerations than sticking to the surface and therefore be preferable. However, slipping down is forbidden because these coefficients are only valid for slipping up and sticking, which have already been established to be optimal.

\begin{figure}[t]
    \centering
    \includegraphics[width=0.45\textwidth]{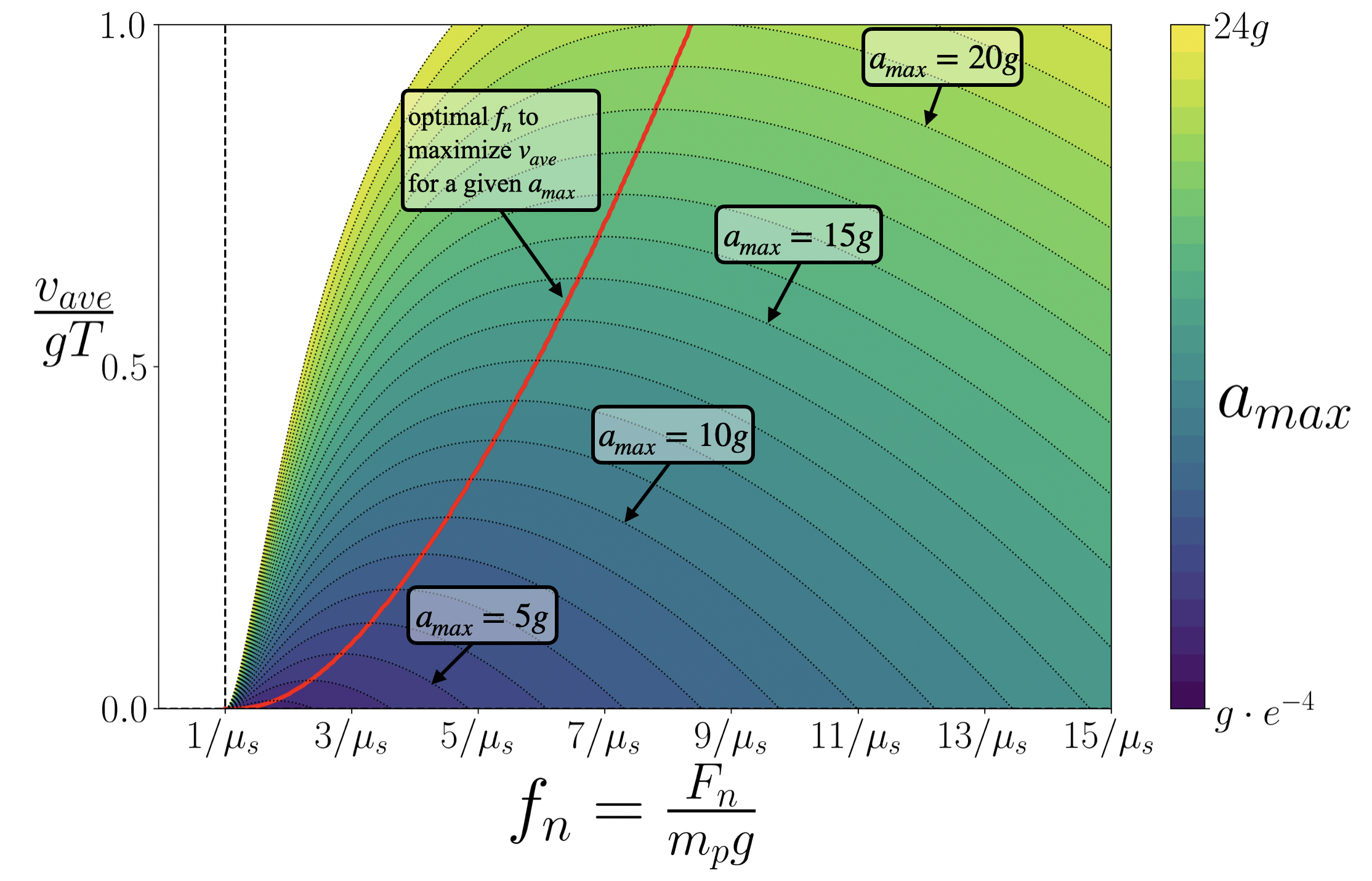}
    \caption{Average normalized velocity versus the normal force per part weight ($f_n$), when $\mu_s = 0.7$ and $\mu_k = 0.6$. 
    Average normalized velocity curves for a given $a_{max}$ are denoted by the dotted lines and several callouts; lighter colors indicate curves with higher maximum accelerations. 
    The red line indicates the value of $f_n$ that maximizes $\frac{v_{ave}}{gT}$ for a given $a_{max}$. 
    Note that this optimal $f_n$ can be an order of magnitude or greater than the force required to statically hold the part.
    The normal force at which a given $a_{max}$ curve intersects the dotted line given by $\frac{v_{ave}}{gT} = 0$ is $f_{n,max}$.}
    \label{figure: v_ave}
    \vspace{\shift}
\end{figure}

It follows from Umbanhowar and Lynch~\cite{umbanhowar2008optimal} that, with $f_n = F_n / (m_P g)$ the normal force per part weight, the optimal surface motion for upward vertical transport is $\ddot{z}_S \colon \left[0,T\right] \rightarrow \left[-a_{max}, a_{max}\right]$ given by:
\begin{equation}
    \ddot{z_S}\colon t \mapsto
    \left\{\begin{aligned} 
        &(\mu_k f_n -1) g, &0 \leq t < T_1 &\quad \text{or} \quad t = T\\
        &-a_{max}, &T_1 \leq t < T_2\\
        &a_{max}, &T_2 \leq t < T
    \end{aligned} \right.
\end{equation}
where
\begin{equation}
\left\{\begin{aligned} 
    T_1 &= T \frac{(\mu_k f_n + 1)}{(\mu_s+\mu_k)f_n}\\
    T_2 &= T_1 + T \frac{(\frac{a_{max}}{g} + \mu_k f_n + 1)(\mu_s f_n -1)}{2\frac{a_{max}}{g}(\mu_s+\mu_k)f_n}
\end{aligned} \right.
\end{equation}
The velocity $\dot{z}_S$ is the integral of $\ddot{z}_S$ with zero mean. The surface motion $z_S$ scales with $T^2$, so higher frequency leads to a smaller surface range of motion.

An example plot of the optimal motion is given in Fig. \ref{figure: optimal motion}.
The three phases are: a sticking phase where the surface and part move upwards, a slipping phase where the surface slips below the part, and another slipping phase where the surface catches up to the part.
In practice, this part motion can be difficult to achieve because it requires the transition at $\frac{t}{T}\in \mathbb{N}$ to happen exactly when the part and surface velocities match up. Operating a little below the slipping limit for $t \in [0, T_1[$ can relax this strict timing requirement.

The corresponding average part velocity~\cite{umbanhowar2008optimal}, normalized by $gT$, is:
\begin{equation}\label{eq: v_ave optimal}
    \frac{v_{ave}}{gT} = \frac{(\mu_s f_n -1)^2((\frac{a_{max}}{g})^2 - (\mu_k f_n + 1)^2)}{4\frac{a_{max}}{g}(\mu_s+\mu_k)^2f_n^2}
\end{equation}
From \eqref{eq: v_ave optimal} we deduce the maximum normal force beyond which upward motion is impossible:
\begin{equation}\label{eq: f_n,supp}
    f_{n,max} = \frac{F_{n,max}}{m_P g} = \frac{1}{\mu_k}\left(\frac{a_{max}}{g} - 1\right)
\end{equation}
This force is the limit where the resulting kinetic friction becomes so large that the surface can no longer slip below the part for $t \in [T_2, T[$: $\ddot{z}_P = \mu_k F_{n,max} / m_P - g = -a_{max}$. 
Hence there is a compromise between squeezing harder to stick more on the way up and not squeezing too hard when dragging the surface back down. 
The optimal force between $1/\mu_s$ and $f_{n,max}$ is the root of a cubic equation given by $\partial v_{ave} / \partial f_n = 0$ and can be numerically estimated.
A graph of the average velocity normalized by $gT$ is given in Fig. \ref{figure: v_ave}, showing that the optimal normal force is large and scales quadratically with $a_{max}$. 
As can be seen in the plot, the optimal normal force can be multiple times that required for static equilibrium~\eqref{eq: F_n lower bound mu_s}, further necessitating high accelerations~\eqref{eq: a_max lower bound}.
This realization led us to use impact-induced accelerations to drive the surface, which can achieve the high accelerations necessary for vertical vibratory transport.
Additionally, impacts can help satisfy the implicit assumption in the optimal acceleration profile that desired accelerations can be instantly reached, in other words, impacts can rapidly achieve the desired accelerations.
Note that we are not directly impacting the part in order to achieve manipulation, as was done by Huang~\cite{huang1997vibratory}.
To make the part go down, one could simply reduce $f_n$ until slipping occurs. Alternatively, applying the same optimization to $-v_{ave}$ yields the surface motion which maximizes the part's downward velocity without changing $f_n$.

\section{DESIGN OF EXPERIMENTAL SETUP}\label{section: design of experimental setup}

We verify the dynamical model by using experimentally measured surface motion to predict part motion. 
This section details the design of a vertical vibratory transport capable device as well as the tools used for gathering data.

\subsection{Physical Build}

\begin{figure}[tb]
    \centering
    \includegraphics[width=0.42\textwidth]{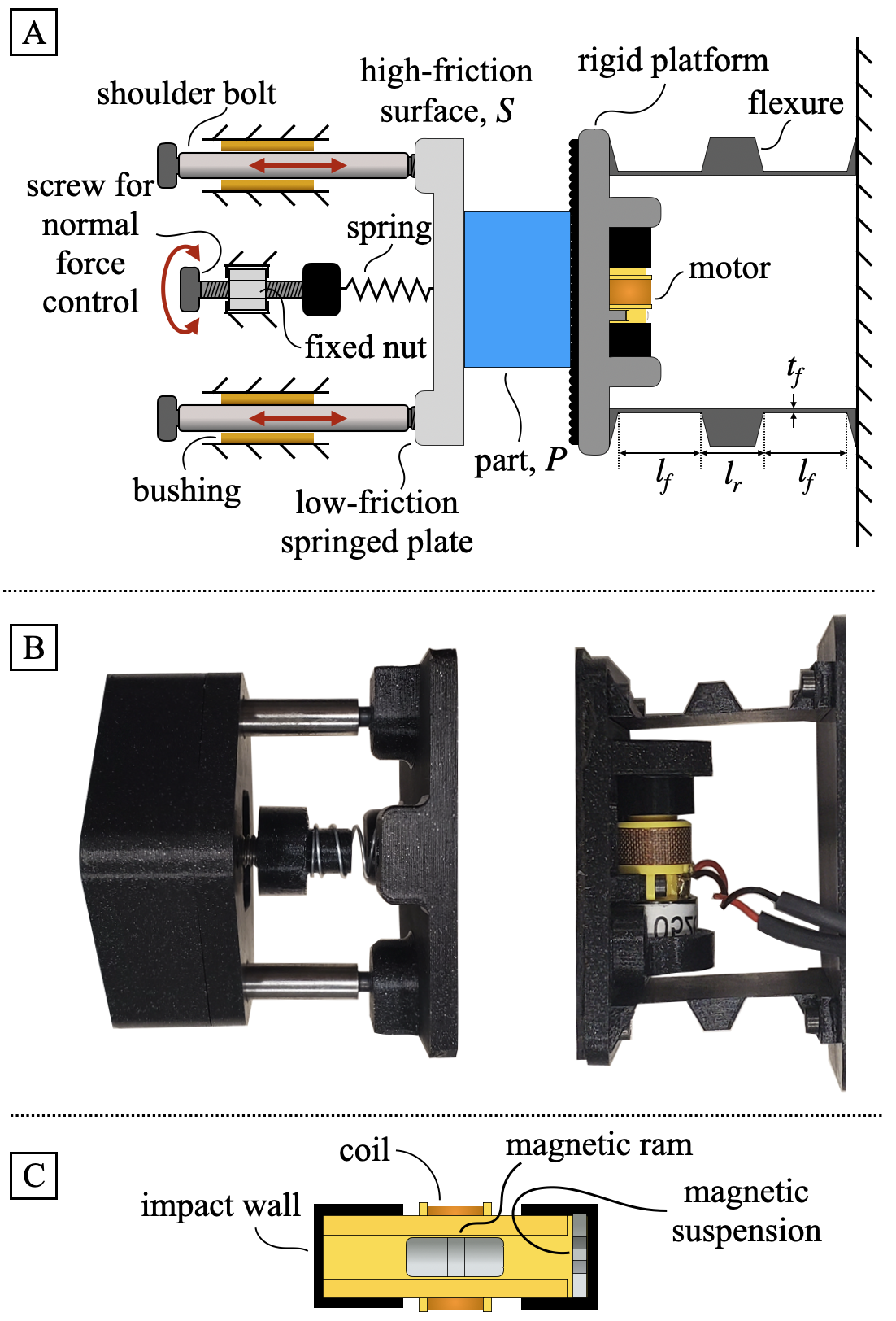}
    \caption{Schematic of experimental setup is shown in (A), with the constructed setup shown in (B) and a section view of the impact motor in (C). 
    (A, B) The vibrating surface $S$ is shown on the right and the constant force springed plate is shown on the left. 
    Movement of the ram inside the motor causes the surface to move. 
    Flexure stiffness and actuator-platform assembly mass were minimized to avoid filtering movement of the ram. 
    A screw and linear spring are used to adjust the normal force applied to $P$.
    Note that springed plate assembly in (B) is viewed from above for clarity, as the shoulder bolts and spring lie in the same plane.
    (C) The magnetic suspension holds the ram at an equilibrium position. 
    The coil can move the ram towards the suspension, which repels the ram, or away from the suspension possibly generating a collision with the impact wall.}
    \label{figure: experimental_setup}
    \vspace{\shift}
\end{figure}

The experimental setup consists of two main components: the oscillating high-friction surface, and constant force low-friction springed plate (Fig. \ref{figure: experimental_setup}A 
and \ref{figure: experimental_setup}B). 
The motor is a Carlton voice-coil haptic actuator from Titan Haptics (section view shown in Fig. \ref{figure: experimental_setup}C and it was selected because of its high bandwidth and ability to rapidly achieve high accelerations in a small form factor.
Electromagnetic forces send the 4 gram magnetic ram either towards the magnetic suspension, which holds the ram at an equilibrium position in the center of the actuator, or the plastic end cap (impact wall).
With sufficient current, the suspension force can be overcome and impacts can be achieved in either direction.
Forces acting on the ram from the magnetic suspension, coil, and actuator structure (i.e., impacts) are applied equally and in the opposite direction to the surface causing it to move.


%
The surface is connected to ground via 3D-printed PLA flexures.
To minimize attenuation of the ram forces on the surface motion, we sought to balance the desired low stiffness of the flexures while ensuring they were strong enough support the actuator-platform assembly, as well as reduce the mass of the entire assembly.
We printed several iterations, varying the length, $l_f$, and thickness, $t_f$, of the flexible portion in addition to the length of the rigid portion, $l_r$.
The selected flexures have the following properties: $l_f = 8 \text{ mm}$, $t_f = 0.3 \text{ mm}$, and $l_r = 9.92 \text{ mm}$.

The low-friction springed plate adjusts the applied normal force by means of a screw and a grounded hex nut. 
Shoulder bolts and bushings were used to minimize motion of the plate along or about axes orthogonal to the screw's axis. 
The spring used was measured to have a stiffness of 0.57 N/mm.
The low-friction surface was smoothed PLA from the bed-facing side of a 3D printed part. 
The coefficient of kinetic friction between the acrylic sheet used in our experiments and this smoothed PLA surface was calculated to be 0.18 and is assumed to be negligible in our model.


\subsection{Surface Motion}


In our analysis, we assume we can prescribe surface motion to match a particular acceleration waveform.
In practice, we can only directly prescribe the driving force acting on the surface, whereas the resulting surface motion is coupled with the motion of the part being picked up, the normal force applied to the part, potential external mechanical stops and springs, and the electrodynamic response of the driving mechanism.
This coupling can be resolved either by using a theoretical model to drive the surface in open-loop, by adding sensors to achieve closed-loop control, or by empirically determining which force waveform to apply.

Additionally, the optimal waveform previously discussed in Section~\ref{section: dynamics} can be difficult to achieve as it requires knowledge of the part mass, the coefficients of friction, and precise timing. 
We instead attempted to achieve the two-stage waveform from Quaid~\cite{quaid1999feeder}, which, although it may result in a slower part velocity, is easier to tune in practice.
We used a voltage-controlled current source to drive the motor and empirically determined a sawtooth current waveform closely achieved the desired low-acceleration upward and large-acceleration downward motion profile.
The drive circuitry was reproduced from previous works~\cite{choi2017grabity, mcmahan2014dynamic}, with the addition of several potentiometers to control the sawtooth waveform's frequency, amplitude, and offset.
A human operator adjusted the knobs based on visual and auditory feedback, and the two-phase stick-slip behavior of the part was later validated using motion tracking data.

\subsection{Data Acquisition}


\begin{figure}[tb]
    \centering
    \includegraphics[width=0.45\textwidth]{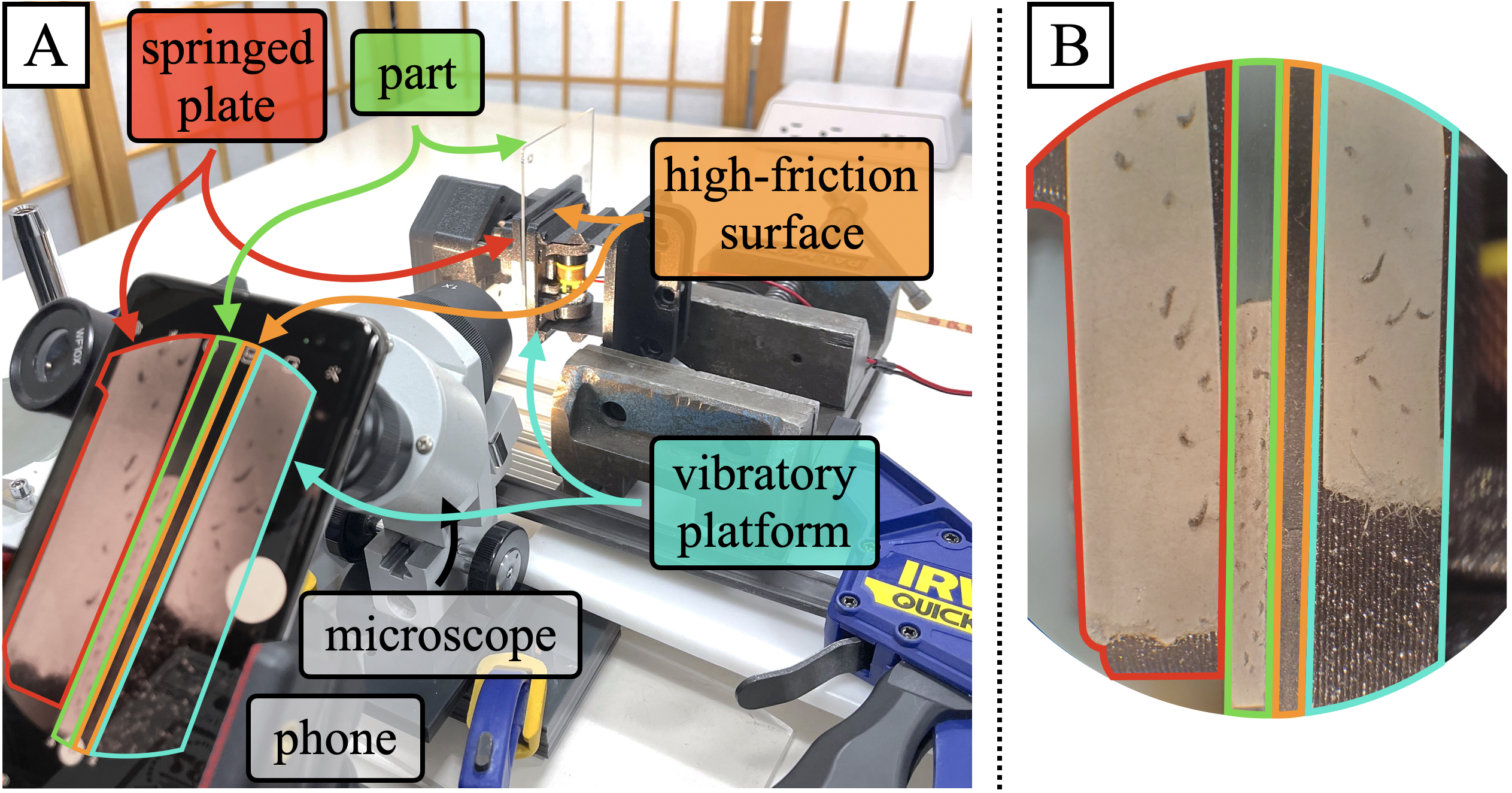}
    \caption{Motion capture setup. 
    (A) The phone camera looks through the lens of a microscope, and observes the motion of the system.
    White packaging labels are marked with dots using a mechanical pencil and affixed to the surface, part, and springed plate; these labels can be seen on the screen of the phone.
    Each of the tracked components are labeled in the physical setup as well as in the view of the phone.
    Movement of the pencil marks are recorded at 960 fps and fed into a motion tracking software to extract position data.
    (B) Close-up of the labels as seen on the phone screen.
    From left to right we have the springed plate, part, high-friction surface, and vibratory platform.}
    \label{figure: recording setup}
    \vspace{\shift}
\end{figure}

We used the setup shown in Fig. \ref{figure: recording setup} to get detailed motion data of the surface and part. 
A phone was secured in a tripod with its camera aligned along the optical axis of a microscope. 
We cut strips off of a white shipping label and affixed them to the surface, part, and springed plate in view of the camera.
A mechanical pencil was then used to mark several dots on the labels. 
Each vertical transport experiment was recorded using the Super Slow-Motion mode of the phone, which can capture 960 fps for 0.4 seconds. 
Videos were then exported to Tracker~\cite{brown2014tracker}, where the points were marked using a combination of autotracking and by hand depending on the degree of motion blur. 
Position data, along with velocity and acceleration data calculated using finite differences, were exported to be used in model validation.

\section{MODEL VALIDATION}\label{section: model validation}

We validated our model by comparing the simulated part position with the experimental part position data. 
Velocity profiles of the surface were computed from the tracked position data using finite differences, and then used as the simulated surface velocity.
We found that the finite difference calculated velocity in Tracker was smoother than the position-differentiated velocity in Simulink/MATLAB, which was much more discretized.
This was important for stabilizing the sticking-slipping transitions that are dependent on the relative velocities.

Ten different transport experiments on a 9 gram acrylic plate were performed across a range of drive frequencies (9.7 to 50.7 Hz), amplitudes (4.7 to 10.6 $V_{p\text{-}p}$), offsets (-2.5 to 2.6 $V$), and normal forces (48 to 206 grams, equivalent to roughly 4 - 16 times $\mu_s m_p g$).
Values were chosen based on empirically observed limits of when transport would and would not occur and limits of the drive circuitry. This allowed us to test the model's capability to reproduce both successful and unsuccessful transport cases.
We used two unknown parameters, $\mu_k$ between the surface and the part, and $F_n$, to fine tune the model both by hand, to narrow in on an acceptable range of a given parameter, and by using a population-based stochastic optimizer (MATLAB's \texttt{particleswarm}) for more precision. 
The optimizer sought to minimize the sum of the per trial mean part position error between the experiment and the model.

\begin{figure}[b]
    \vspace{\shift}
    \centering
    \includegraphics[width=0.5\textwidth]{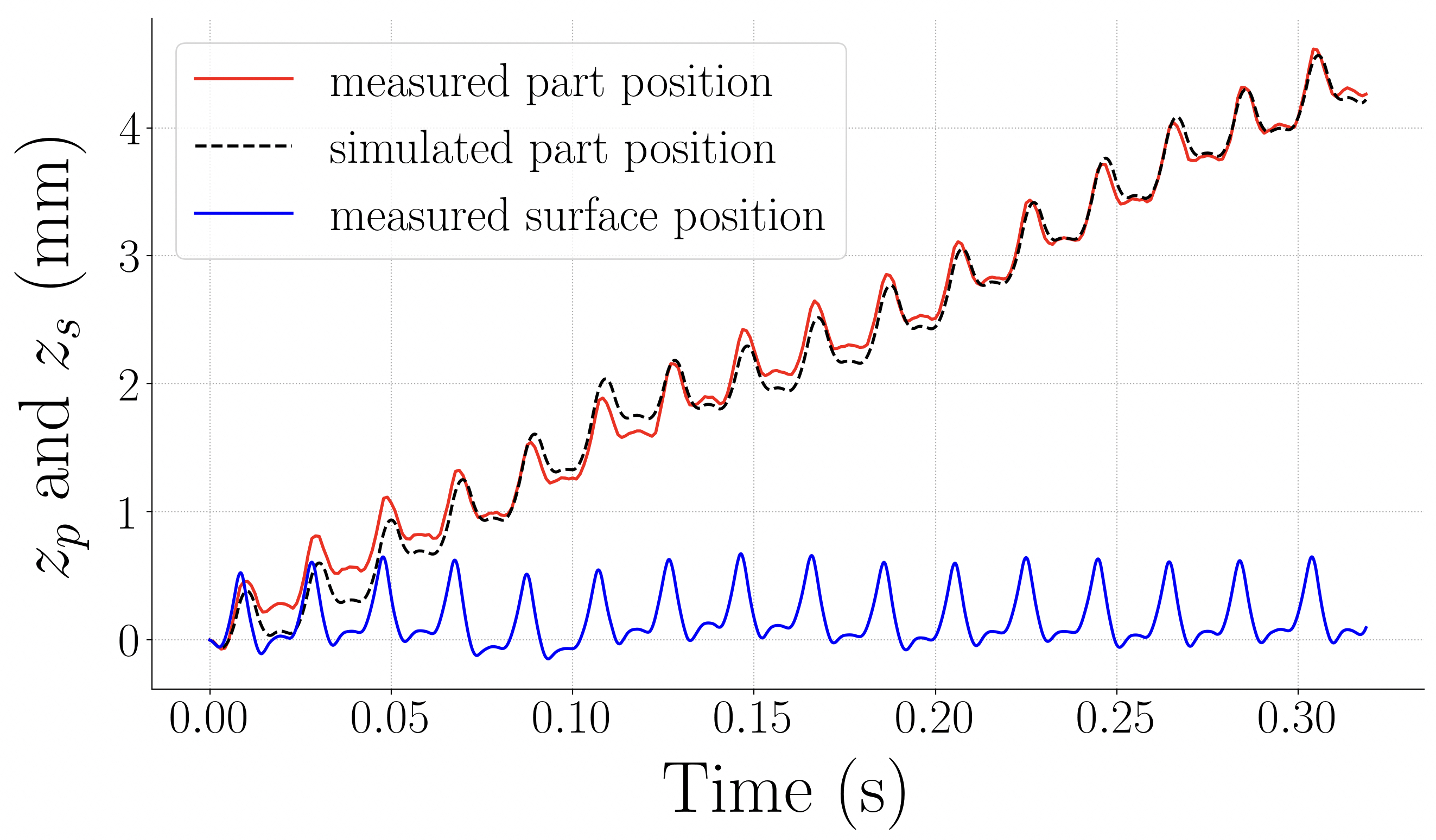}
    \caption{Experimentally measured part position is shown as the black dotted line, with the simulated part position shown in red. The surface motion is shown in blue.}
    \label{figure: model validation}
\end{figure}
The average part position error was 0.16 mm, or $12\%$ when normalizing by each trial's maximum experimentally measured part position. 
The fit normal force deviated from its measured value by a mean of 34\%. 
A majority of the fit normal forces were higher than their experimentally measured counterparts, which we believe is due to the unmodeled friction forces from the springed plate on the part.
We also find the kinetic coefficient of friction between the surface and part to be $\mu_k=0.62$ (measured $\mu_k = 0.59$); a measured $\mu_s=0.72$ was used in the model.
A sample trial is shown in Fig. \ref{figure: model validation}.
The surface position is similar in shape to the quadratic curve in Quaid~\cite{quaid1999feeder}.
Given the strong agreement between the model and experimental data across all trials, we believe that Coulomb friction and the presented model are sufficient to capture the system behavior.

\section{PROOF-OF-CONCEPT GRIPPER}\label{section: gripper}

\begin{figure}[t]
    \centering
    \includegraphics[width=0.45\textwidth]{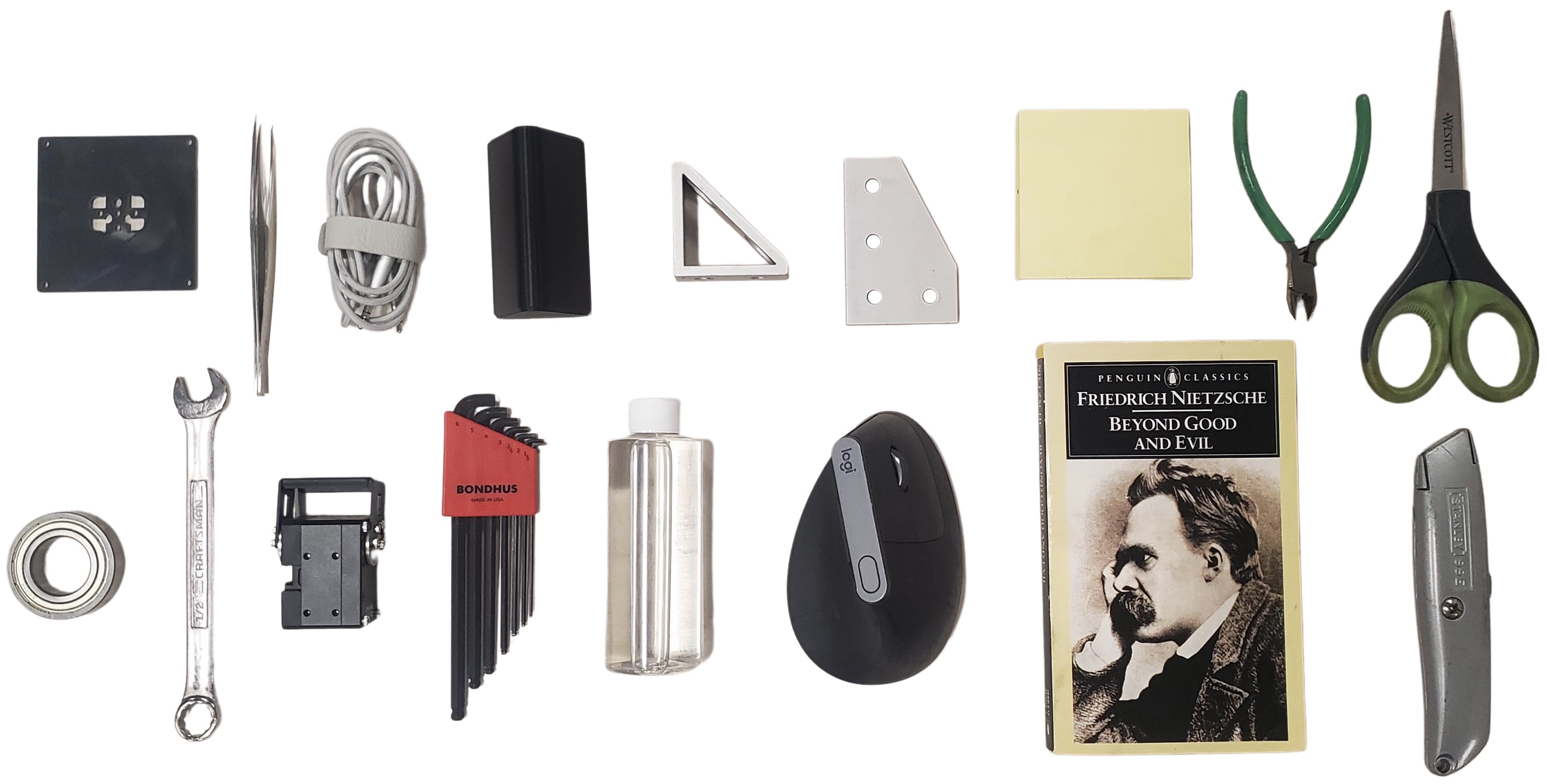}
    \caption{Parts transported by the gripper. From left to right the parts are: acrylic plate, tweezers, bundled braided cable, PLA triangular prism, 80/20 corner bracket and plate, Post-Its, pliers, scissors, bearing, wrench, Dynamixel, hex keys, bottle of fluid, computer mouse, soft cover book, box cutter. The parts ranged in mass from 17 - 169 grams (from left to right, and top to bottom).}
    \label{figure: grasped parts}
    \vspace{\shift}
\end{figure}

A proof-of-concept parallel jaw gripper was designed where each jaw was outfitted with a vibrating surface as shown in Fig. \ref{figure: cover photo}. 
We tested the device's vertical transport ability across a range of different parts as shown in Fig. \ref{figure: grasped parts}.
The gripper was able to transport each part, with operator-tuned sawtooth waveform parameters.
We refer the readers to the accompanying video for transportation of these parts.

From our testing we found that compliant parts as well as light parts were difficult to transport. 
For the former, the surface could neither accelerate up nor slip behind the part. 
Regarding the latter, slight misalignment between the two vibrating surfaces needed to be corrected by applied normal forces, and since we know from Fig. \ref{figure: v_ave} that excessively large normal forces prevent any part motion, we hypothesize that the normal forces required to correct alignment easily exceeded the motion threshold because of their small mass.

\section{DISCUSSION}\label{section: discussion}


\subsection{Direct Drive}
The actuator used was able to quickly supply sufficiently large accelerations in a small form factor, however, we were also curious about a direct drive solution.
Such a device could allow for easier waveform shaping, and possibly deliver even higher accelerations without needing impacts.
To investigate this, we deconstructed an ND91-4$\Omega$ speaker into its moving coil and permanent magnet and attached a high friction surface to the coil.
Despite its higher acceleration capability compared with the Carlton motor, we found it to be quite difficult to achieve transport across  various waveforms and normal forces.
The conditions for achieving transport were significantly more forgiving with the Carlton motor.

Given this, we decided to add impact plates, i.e., hardstops, to contact a protrusion fixed to the moving coil to rapidly decelerate it at the end of the sticking phase.
While this did achieve transport in specific cases, agreement between the model's fit and the experimental data was poor. 
We hypothesize that this disagreement is due to inconsistent normal forces.
The larger vertical range of motion of the platform also results in more horizontal travel because of the flexure constraint.
This causes the springed plate to experience to be compressed / relaxed to a larger extent, violating the constant normal force assumption in our modeling.

\subsection{Limitations}

As noted by Umbanhowar and Lynch~\cite{umbanhowar2008optimal}, the addition of out-of-plane vibrations can greatly increase the optimal part velocity.
While we chose to assume idealized 1D in-plane vibrations for the model, our flexures do not prescribe completely linear motion.
The tracking data does show some horizontal motion of the surface, part, and springed plate, however, it is an order of magnitude smaller than that of the vertical motion.
Given that, any additional compression of the spring is taken to be a result of horizontal inertial forces from the surface and part, implying a constant normal force.

Finding the ideal driving waveform was nontrivial with our setup; the dynamics of the impact motor are difficult to predict and therefore specific acceleration profiles were determined empirically.
Even then, we were unable to achieve the precise sawtooth velocity profile discussed by Quaid~\cite{quaid1999feeder}, let alone verify the optimal surface motion discussed in Section~\ref{section: dynamics}.
Additionally, the plane-on-plane contact between the part, surface, and springed plate enforced a difficult alignment constraint for achieving evenly distributed friction forces.
A roller with point contact could have been used as in Nahum~\cite{nahum2022robotic}, as well as compliance in the flexure mount similar to the hand by Cai~\cite{cai2023hand}.
Misalignment, as well as differences in the electrical characteristics of each motor and tilting of the gripper could also have caused the occasional non-synchronized motion of the surfaces observable in the video attachment.
Regardless, these imprecisions did not seem to significantly impact the performance of the device against the springed plate or when used in the gripper.


\section{CONCLUSIONS AND FUTURE WORK}\label{section: conclusion and future work}


Our research demonstrated a method for achieving vertical transport of grasped parts using vibratory fingertip motions to induce periodic sticking and slipping phases of motion. 
A theoretical model of the process was proposed and validated. 
It was used to show analytically that large normal forces and rapid achievement of large accelerations are key to realizing vertical transport. 
We achieved rapid onset of gripper pad motion by using impact-induced accelerations. 
Experiments confirmed our analytical model and a gripper demonstrated the ability of our method to vertically transport a wide variety of parts. 
We envision this work being used for robotic in-hand planar manipulation using multiple vibration sources and closed-loop grasp force modulation.
A gripper with two planar manipulation capable surfaces and individually controllable finger joints could easily perform holonomic spatial manipulation in a mechanically simple package.
Furthermore, in-hand manipulation during a power grasp could be achieved by placing multiple of these surfaces along each phalanx of a multi-fingered, multi-link hand.

\addtolength{\textheight}{-18cm}   




\bibliographystyle{ieeetr}
\bibliography{root}

\end{document}